\documentclass[twocolumn,a4paper]{article}
\usepackage[utf8]{inputenc}

\usepackage[a4paper,margin=0.9in]{geometry}

\usepackage{titlesec}
\titleformat{\section}{\normalfont\large\bfseries}{\thesection}{0pt}{}
\titleformat{\subsection}{\normalfont\bfseries}{\thesubsection}{0pt}{}

\usepackage[superscript,biblabel]{cite}

\usepackage{graphicx}

\usepackage[breaklinks=true]{hyperref}
\usepackage{breakcites}

\usepackage{array}

\title{Characterizing Political Fake News in Twitter by its Meta-Data}
\author{Julio Amador~D\'iaz~L\'opez\and Axel Oehmichen \and Miguel Molina-Solana}
\date{\textit{( j.amador, axelfrancois.oehmichen11, mmolinas@imperial.ac.uk )} \\ Imperial College London}

\begin{document}

\twocolumn[
  \begin{@twocolumnfalse}
    \maketitle
    \begin{abstract} 
This article presents a preliminary approach towards characterizing political fake news on Twitter through the analysis of their meta-data. In particular, we focus on more than 1.5M tweets collected on the day of the election of Donald Trump as 45th president of the United States of America. We use the meta-data embedded within those tweets in order to look for differences between tweets containing fake news and tweets not containing them. Specifically, we perform our analysis only on tweets that went viral, by studying proxies for users' exposure to the tweets, by characterizing accounts spreading fake news, and by looking at their polarization. We found significant differences on the distribution of followers, the number of URLs on tweets, and the verification of the users.

\end{abstract}
  \end{@twocolumnfalse}
  ]

\section*{Introduction}

While \textit{fake news}, understood as deliberately misleading pieces of information, have existed since long ago (e.g.\ it is not unusual to receive news falsely claiming the death of a celebrity), the term reached the mainstream, particularly so in politics, during the 2016 presidential election in the United States~\cite{Guardian2dec}. Since then, governments 
and corporations alike (e.g.\ Google~\cite{GoogleFN} and Facebook~\cite{FacebookFN}) have begun efforts to tackle fake news as they can affect political decisions~\cite{NYtimes}. Yet, the ability to define, identify and stop fake news from spreading is limited. 

Since the Obama campaign in 2008, social media has been pervasive in the political arena in the United States. Studies report that up to 62\% of American adults receive their news from social media~\cite{Gottfried2016}. The wide use of platforms such as Twitter and Facebook has facilitated the diffusion of fake news by simplifying the process of receiving content with no significant third party filtering, fact-checking or editorial judgement. Such characteristics make these platforms suitable means for sharing news that, disguised as legit ones, try to confuse readers.

Such use and their prominent rise has been confirmed by Craig Silverman, a Canadian journalist who is a prominent figure on fake news \cite{Silverman2015}: ``In the final three months of the US presidential campaign, the top-performing fake election news stories on Facebook generated more engagement than the top stories from major news outlet''. 

Our current research hence departs from the assumption that social media is a conduit for \textit{fake news} and asks the question of whether \textit{fake news} (as \textit{spam} was some years ago) can be identified, modelled and eventually blocked. In order to do so, we use a sample of more that 1.5M tweets collected on November 8th 2016 ---election day in the United States--- with the goal of identifying features that tweets containing fake news are likely to have. As such, our paper aims to provide a preliminary characterization of fake news in Twitter by looking into meta-data embedded in tweets. Considering meta-data as a relevant factor of analysis is in line with findings reported by Morris et al.~\cite{Morris2012}. We argue that understanding differences between tweets containing fake news and regular tweets will allow researchers to design mechanisms to block fake news in Twitter.

Specifically, our goals are: 1) compare the characteristics of tweets labelled as containing fake news to tweets labelled as not containing them, 2) characterize, through their meta-data, viral tweets containing fake news and the accounts from which they originated, and 3) determine the extent to which tweets containing fake news expressed polarized political views. 

For our study, we used the number of retweets to single-out those that went viral within our sample. Tweets within that subset (viral tweets hereafter) are varied and relate to different topics. We consider that a tweet contains fake news if its text falls within any of the following categories described by Rubin et al.~\cite{Rubin2015} (see next section for the details of such categories): serious fabrication, large-scale hoaxes, jokes taken at face value, slanted reporting of real facts and stories where the truth is contentious. The dataset~\cite{fakenews_1day_dataset}, manually labelled by an expert, has been publicly released and is available to researchers and interested parties.

From our results, the following main observations can be made:
\begin{itemize}
\item Distribution in the number of retweets, favourites and hashtags in tweets containing fake news are not significantly different from their counterparts in tweets not containing fake news.
\item Accounts generating fake news are comparatively more unverified that accounts not producing fake news.
\item There are significant differences in both the number of friends and followers of the accounts creating tweets with fake news when compared with accounts not generating them.
\item There are no significant differences in the number of media elements, but there are indications that the number of URLs it is indeed different.
\end{itemize}



Our findings resonate with similar work done on fake news such as the one from Allcot and Gentzkow~\cite{Allcot2017}. Therefore, even if our study is a preliminary attempt at characterizing fake news on Twitter using only their meta-data, our results provide external validity to previous research. Moreover, our work not only stresses the importance of using meta-data, but also underscores which parameters may be useful to identify fake news on Twitter.  

The rest of the paper is organized as follows. The next section briefly discusses where this work is located within the literature on fake news and contextualizes the type of fake news we are studying. Then, we present our hypotheses, the data, and the methodology we follow. Finally, we present our findings, conclusions of this study, and future lines of work.

\section*{Defining Fake news}

Our research is connected to different strands of academic knowledge related to the phenomenon of fake news. In relation to Computer Science, a recent survey by Conroy and colleagues~\cite{Conroy2015} identifies two popular approaches to single-out fake news. On the one hand, the authors pointed to linguistic approaches consisting in using text, its linguistic characteristics and machine learning techniques to automatically flag fake news. On the other, these researchers underscored the use of network approaches, which make use of network characteristics and meta-data, to identify fake news. 

With respect to social sciences, efforts from psychology, political science and sociology, have been dedicated to understand why people consume and/or believe misinformation~\cite{Pennycook2017,Flynn2017,Polage2012,Swire2017}. Most of these studies consistently reported that psychological biases such as \emph{priming effects} and \emph{confirmation bias} play an important role in people ability to discern misinformation. 

In relation to the production and distribution of fake news, a recent paper in the field of Economics~\cite{Allcot2017} found that most fake news sites use names that resemble those of legitimate organizations, and that sites supplying fake news tend to be short-lived. These authors also noticed that fake news items are more likely shared than legitimate articles coming from trusted sources, and they tend to exhibit a larger level of polarization.

The conceptual issue of how to define fake news is a serious and unresolved issue. As the focus of our work is not attempting to offer light on this, we will rely on work by other authors to describe what we consider as fake news. In particular, we use the categorization provided by Rubin et al.~\cite{Rubin2015}. The five categories they described, together with illustrative examples from our dataset, are as follows:

\begin{enumerate}
\item {\bf Serious fabrication.}
These are news stories created entirely to deceive readers. During the 2016 US presidential election there were plenty of examples of this (e.g.\ claiming a celebrity has endorsed Donald Trump when that was not the case). For instance: [@JebBush - {\it Maybe Donald negotiated a deal with his buddy @HillaryClinton. Continuing this path will put her in the White House. \url{https://t.co/AlvByiSrMn}}]

\item {\bf Large-scale hoaxes.}
Deceptions that are then reported in good faith by reputable sources. A recent example would be the story that the founder of Corona beer made everyone in his home village a millionaire in his will. For instance: [@FullFrontalSamB - {\it Unfortunately Melania copied HER ballot from Michelle so... Donald just voted for Hillary. \#ElectionDay \url{https://t.co/x2ZimtFxyl}}]

\item {\bf Jokes taken at face value.}
Humour sites such as the Onion or Daily Mash present fake news stories in order to satirise the media. Issues can arise when readers see the story out of context and share it with others. For instance: [@BBCTaster - {\it BREAKING NEWS: If you face-swap @realDonaldTrump with @MayorofLondon you get Owen Wilson. \url{https://t.co/YY8a20wQVP}}]

\item {\bf Slanted reporting of real facts.}
Selectively-chosen but truthful elements of a story put together to serve an agenda. One of the most prevalent examples of this is the well-known problems of voting machine faults. For instance: [@NeilTurner\_ - {\it @realDonaldTrump Trump predicted it. \#BrusselsAttack \url{https://t.co/BM3UxA7heR}}]

\item {\bf Stories where the `truth' is contentious.}
On issues where ideologies or opinions clash ---for example, territorial conflicts--- there is sometimes no established baseline for truth. Reporters may be unconsciously partisan, or perceived as such. For instance: [@FoxNews - {\it Report: @HillaryClinton's plan would raise taxes \$1.3T/10 years. \url{https://t.co/Dh1tWM4FAP}}]

\end{enumerate}


\section*{Research Hypotheses}

Previous works on the area (presented in the section above) suggest that there may be important determinants for the adoption and diffusion of fake news. Our hypotheses builds on them and identifies three important dimensions that may help distinguishing fake news from legit information:
\begin{enumerate}
\item {\bf Exposure.} Given that psychological effects such as priming and confirmation biases are likely to increase the probability an individual believes in a certain piece of information, we believe exposure to misinformation is an important determinant of a fake news distribution strategy.
\item {\bf Characterization.} Given that distributors of fake news may want to simulate legitimate information outlets, we believe it is important to analyse specific features that may help a fake news outlet `disguise' as a legit one.
\item {\bf Polarization.} Given that fake news outlets are more likely to attract attention with polarizing content (See  \cite{Swire2017}), we believe the level of polarization is an important determinant of a fake news distribution strategy. 
\end{enumerate}

Taking those three dimensions into account, we propose the following hypotheses about the features that we believe can help to identify tweets containing fake news from those not containing them. They will be later tested over our collected dataset.


\noindent\textbf{\emph{Exposure.}} \begin{description}
	\item[H1A:] The average number of retweets of a viral tweet containing fake news is larger than that of viral tweets not containing them. 
	\item[H1B:] The average number of hashtags and user mentions in viral tweets with fake news is larger than that of viral tweets with no fake news in them. 
\end{description}
\textbf{\emph{Characterization.}}  \begin{description}
	\item[H2A:] Viral tweets containing fake news have a larger number of URLs. 
	\item[H2B:] Creation date of an account generating tweets with fake news is more recent that those accounts tweeting non-fake news content.
	\item[H2C:] The rate of friends/followers of accounts tweeting fake news is larger than the rate of those creating tweets without them.
\end{description}
\textbf{\emph{Polarization.}}  \begin{description}
	\item[H3:] Viral tweets containing fake news are slanted towards one candidate.
\end{description}

\section*{Data and Methodology}

For this study, we collected publicly available tweets using Twitter's public API. Given the nature of the data, it is important to emphasize that such tweets are subject to Twitter's terms and conditions which indicate that users consent to the collection, transfer, manipulation, storage, and disclosure of data. Therefore, we do not expect ethical, legal, or social implications from the usage of the tweets. Our data was collected using search terms related to the presidential election held in the United States on November 8th 2016. Particularly, we queried Twitter's streaming API, more precisely the \textit{filter endpoint} of the streaming API, using the following hashtags and user handles: \texttt{\#MyVote2016}, \texttt{\#ElectionDay}, \texttt{\#electionnight}, \texttt{@realDonaldTrump} and \texttt{@HillaryClinton}. The data collection ran for just one day (Nov 8th 2016).

One straightforward way of sharing information on Twitter is by using the retweet functionality, which enables a user to share a exact copy of a tweet with his followers. Among the reasons for retweeting, Body et al.~\cite{Boyd2010} reported the will to: 1) spread tweets to a new audience, 2) to show one’s role as a listener, and 3) to agree with someone or validate the thoughts of others. As indicated, our initial interest is to characterize tweets containing fake news that went viral (as they are the most harmful ones, as they reach a wider audience), and understand how it differs from other viral tweets (that do not contain fake news). For our study, we consider that a tweet went viral if it was retweeted more than 1000 times. 

Once we have the dataset of viral tweets, we eliminated duplicates (some of the tweets were collected several times because they had several handles) and an expert manually inspected the text field within the tweets to label them as containing fake news, or not containing them (according to the characterization presented before). This annotated dataset~\cite{fakenews_1day_dataset} is publicly available and can be freely reused.

Finally, we use the following fields within tweets (from the ones returned by Twitter's API) to compare their distributions and look for differences between \emph{viral tweets containing fake news} and \emph{viral tweets not containing fake news}: 

\begin{itemize}
\item \emph{Exposure:} \texttt{created\_at}, \texttt{retweet\_count}, \texttt{favourites\_count} and \texttt{hashtags}.
\item \emph{Characterization.} \texttt{screen\_name}, \texttt{verified}, \texttt{urls}, \texttt{followers\_count}, \texttt{friends\_count} and \texttt{media}.
\item \emph{Polarization.} \texttt{text} and \texttt{hashtags}.
\end{itemize}

In the following section, we provide graphical descriptions of the distribution of each of the identified attributes for the two sets of tweets (those labelled as containing fake news and those labelled as not containing them). Where appropriate, we normalized and/or took logarithms of the data for better representation. To gain a better understanding of the significance of those differences, we use the Kolmogorov-Smirnov test with the null hypothesis that both distributions are equal.

\section*{Results}

The sample collected consisted on 1 785 855 tweets published by 848 196 different users. Within our sample, we identified 1327 tweets that went viral (retweeted more than 1000 times by the 8th of November 2016) produced by 643 users. Such small subset of viral tweets were retweeted on 290 841 occasions in the observed time-window. 

The 1327 `viral' tweets were manually annotated as containing fake news or not. The annotation was carried out by a single person in order to obtain a consistent annotation throughout the dataset. Out of those 1327 tweets, we identified 136 as potentially containing fake news (according to the categories previously described), and the rest were classified as `non containing fake news'. Note that the categorization is far from being perfect given the ambiguity of fake news themselves and human judgement involved in the process of categorization. Because of this, we do not claim that this dataset can be considered a ground truth.

The following results detail characteristics of these tweets along the previously mentioned dimensions.
Table~\ref{meanDifferences} reports the actual differences (together with their associated p-values) of the distributions of viral tweets containing fake news and viral tweets not containing them for every variable considered.


{
\begin{table}[htbp]
\begin{center}
\begin{tabular}{| l | c c |}\hline
\multicolumn{3}{| c |} {Kolmogorov-Smirnov test}\\ \hline \hline
feature & {difference} & {p-value} \\ \hline
Followers & 0.2357 & 2.6E-6\\
Friends & 0.1747 & 0.0012 \\
URLs& 0.1285 & 0.0358\\
Favourites & 0.1218 & 0.0535\\
Mentions & 0.1135 & 0.0862\\
Media & 0.0948 & 0.2231\\
Retweets & 0.0609 & 0.7560\\
Hashtags  & 0.0350 & 0.9983\\
\hline
\end{tabular}
\caption{For each one of the selected features, the table shows the difference between the set of tweets containing fake news and those non containing them, and the associated p-value (applying a Kolmogorov-Smirnov test). The null hypothesis is that both distributions are equal (two sided). Results are ordered by decreasing p-value.}
\label{meanDifferences}
\end{center}
\end{table}
}


\subsection*{Exposure}

\begin{figure}[!ht]
 \centering
  \includegraphics[width=\columnwidth, keepaspectratio]{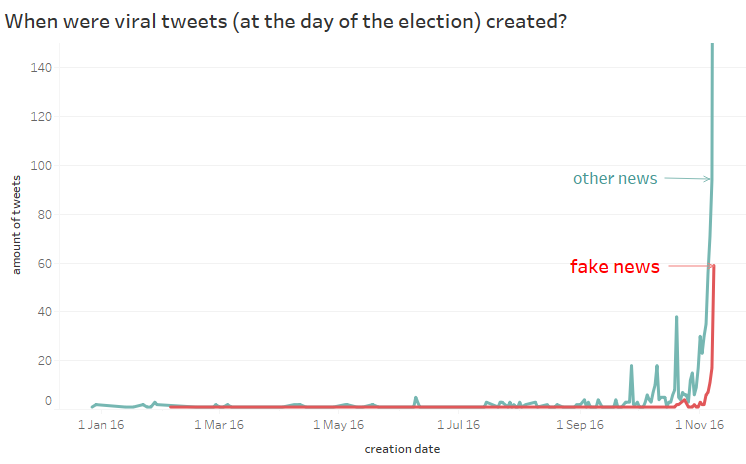}
  \caption{Distribution of the date of creation of the tweets that were viral on November 8th. For clarity, the image only shows the year 2016, and no more than 150 tweets per day.}
  \label{fig:timeline}
\end{figure}

Figure~\ref{fig:timeline} shows that, in contrast to other kinds of viral tweets, those containing fake news were created more recently. As such, Twitter users were exposed to fake news related to the election for a shorter period of time.

\begin{figure}[!ht]
 \centering
  \includegraphics[width=\columnwidth, keepaspectratio]{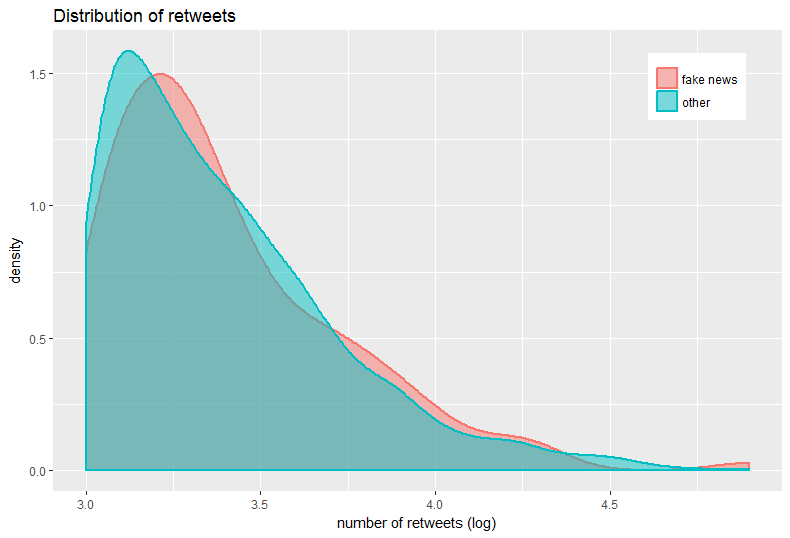}
  \caption{Density distributions of achieved retweets for tweets in our dataset 1)containing fake news and 2)not containing them. No differences are apparent.}
  \label{fig:retweets}
\end{figure}

However, in terms of retweets, Figure~\ref{fig:retweets} shows no apparent difference between containing fake news or not containing them. That is confirmed by the Kolmogorov-Smirnoff test, which does not discard the hypothesis that the associated distributions are equal.

\begin{figure}[!ht]
 \centering
  \includegraphics[width=\columnwidth, keepaspectratio]{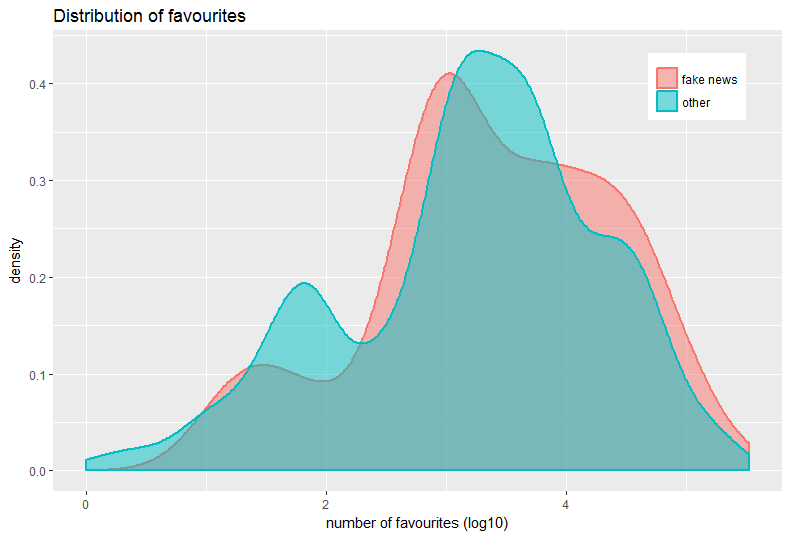}
  \caption{Density distributions of the number of favourites that the user generating the tweet has. The differences are not statistically significant.}
  \label{fig:favourites}
\end{figure}

\begin{figure*}[!ht]
  \includegraphics[width=0.99\textwidth]{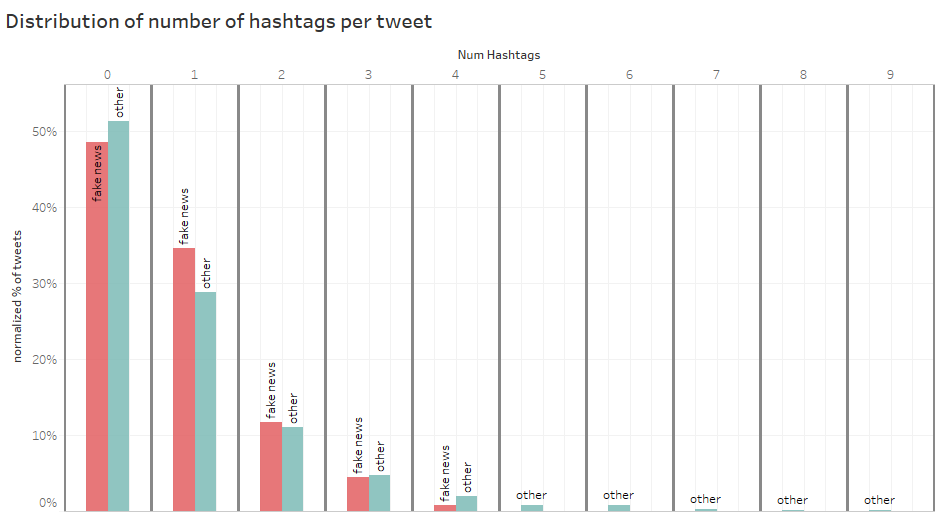}
  \caption{Distribution of the number of hashtags used in tweets labelled as containing fake news and those labelled as not containing them.}
  \label{fig:hashtags}
\end{figure*}

In relation to the number of favourites, users that generated at least a viral tweet containing fake news appear to have, on average, less favourites than users that do not generate them. Figure~\ref{fig:favourites} shows the distribution of favourites. Despite the apparent visual differences, the difference are not statistically significant.

Finally, the number of hashtags used in viral fake news appears to be larger than those in other viral tweets. Figure~\ref{fig:hashtags} shows the density distribution of the number of hashtags used. However, once again, we were not able to find any statistical difference between the average number of hashtags in a viral tweet and the average number of hashtags in viral fake news.

\subsection*{Characterization}

\begin{figure}[!ht]
 \centering
  \includegraphics[width=\columnwidth, keepaspectratio]{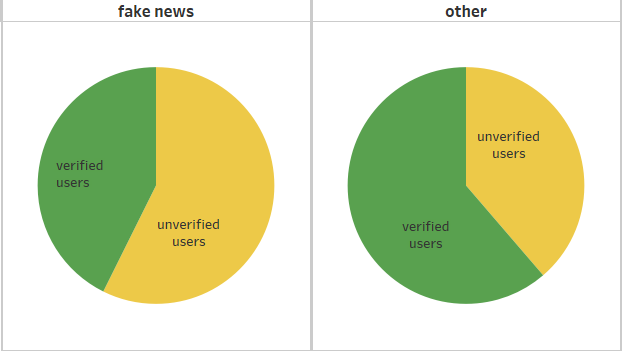}
  \caption{Tweets labelled as containing fake news mostly come from non-verified users. This contrasts with the opposite pattern for tweets non containing them (which mostly originate from verified accounts).}
  \label{fig:verifiedFN}
\end{figure}

We found that 82 users within our sample were spreading fake news (i.e.\ they produced at least one tweet which was labelled as fake news). Out of those, 34 had verified accounts, and the rest were unverified. From the 48 unverified accounts, 6 have been suspended by Twitter at the date of writing, 3 tried to imitate legitimate accounts of others, and 4 accounts have been already deleted. Figure~\ref{fig:verifiedFN} shows the proportion of verified accounts to unverified accounts for viral tweets (containing fake news vs. not containing fake news). From the chart, it is clear that there is a higher chance of fake news coming from unverified accounts.

\begin{figure}[!ht]
 \centering
  \includegraphics[width=\columnwidth, keepaspectratio]{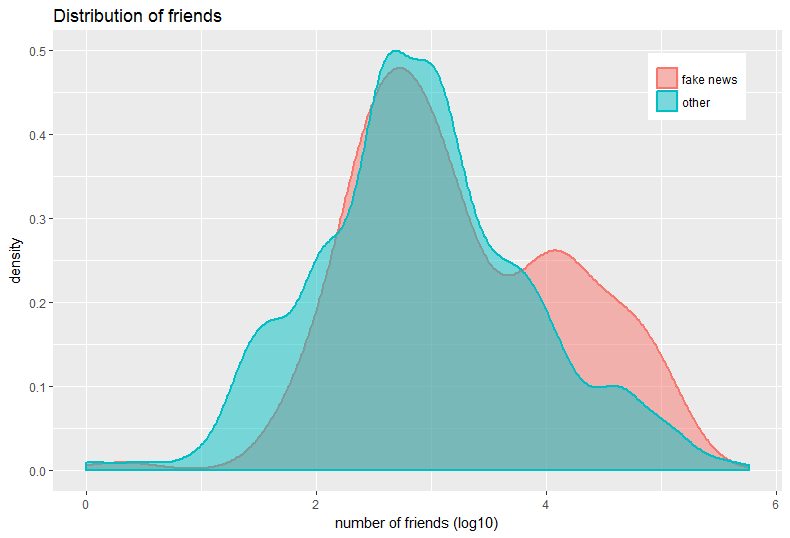}
  \caption{Density distributions (for tweets labelled as containing fake news, and tweets labelled as not containing them) of the number of friends that the user generating the tweet has. Difference is statistically significant.}
  \label{fig:friends}
\end{figure}

Turning to friends, accounts distributing fake news appear to have, on average, the same number of friends than those distributing tweets with no fake news. However, the density distribution of friends from the accounts (Figure~\ref{fig:friends}) shows that there is indeed a statistically significant difference in their distributions.

\begin{figure}[!ht]
 \centering
  \includegraphics[width=\columnwidth, keepaspectratio]{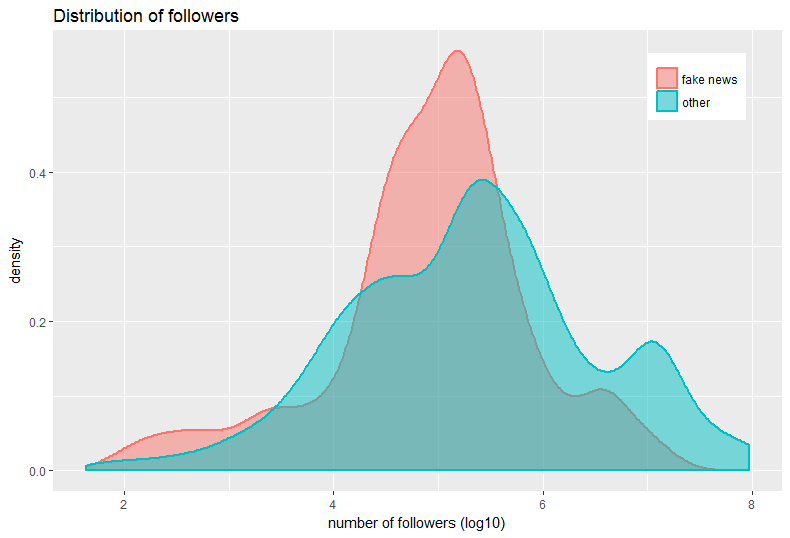}
  \caption{Density distributions of the number of followers that the accounts generating viral tweets (within our sample) have. Accounts producing fake news have a narrower window of followers.}
  \label{fig:followers}
\end{figure}

\begin{figure}[!ht]
 \centering
  \includegraphics[width=\columnwidth, keepaspectratio]{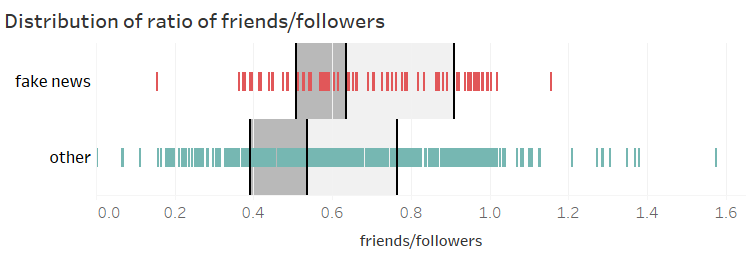}
  \caption{Density distribution of friends/followers ratio, showing quartiles. Accounts that generate fake news tend to have a higher ratio value.}
  \label{fig:ratioFriendsFollowers}
\end{figure}

\begin{figure}[!ht]
 \centering
  \includegraphics[width=\columnwidth, keepaspectratio]{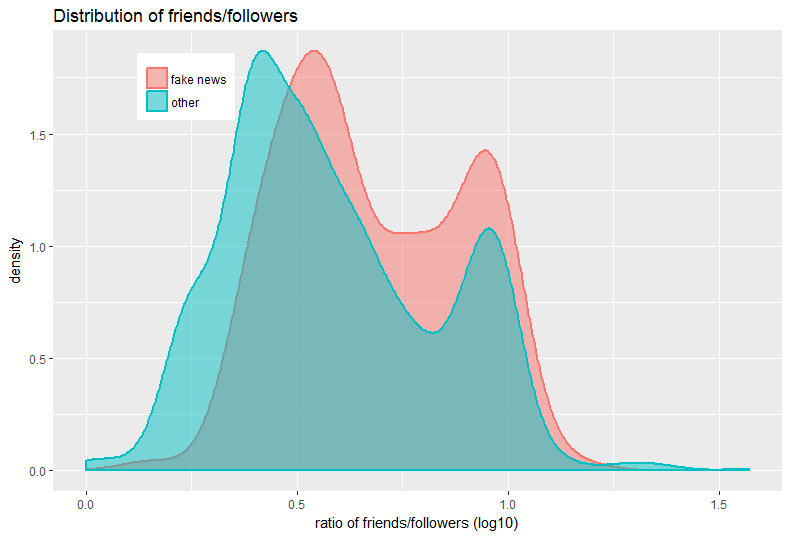}
  \caption{Density distribution of friends/followers ratio. Note that they do not follow a normal distribution. A higher friends/followers ratio exists for accounts that has at least produced a tweet labelled as containing fake news.}
  \label{fig:ratioFriendsFollowersDENSITY}
\end{figure}

If we take into consideration the number of followers, accounts generating viral tweets with fake news do have a very different distribution on this dimension, compared to those accounts generating viral tweets with no fake news (see Figure~\ref{fig:followers}). In fact, such differences are statistically significant. 

A useful representation for friends and followers is the ratio between friends/followers. Figures~\ref{fig:ratioFriendsFollowers} and \ref{fig:ratioFriendsFollowersDENSITY} show this representation. Notice that accounts spreading viral tweets with fake news have, on average, a larger ratio of friends/followers. The distribution of those accounts not generating fake news is more evenly distributed.

\begin{figure}[!ht]
 \centering
  \includegraphics[width=\columnwidth, keepaspectratio]{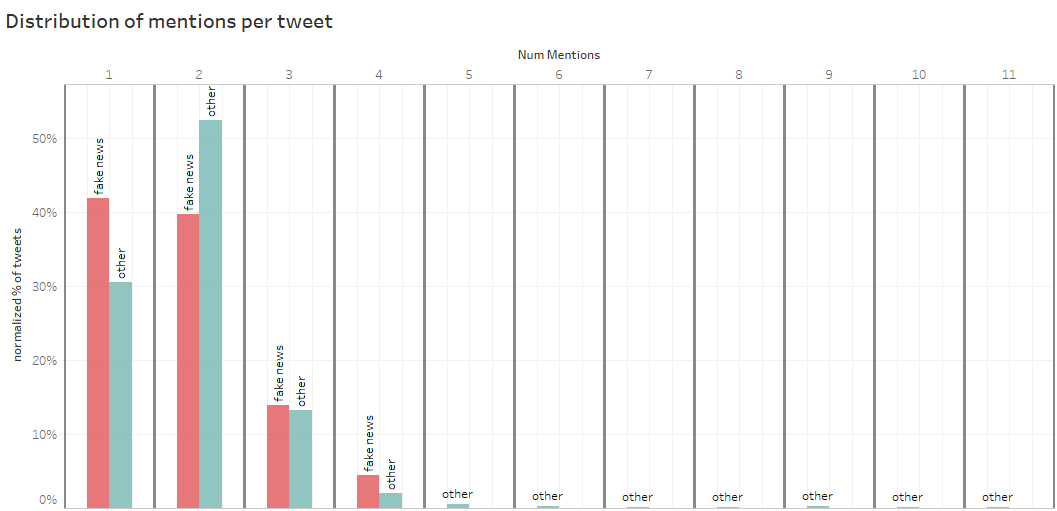}
  \caption{Number of mentions within tweets labelled as containing fake news and tweets not containing them. There is almost a similar distribution of 1 and 2 mentions for tweets containing fake news. This contrasts with tweets not containing fake news, in which 2 mentions is much more common.}
  \label{fig:mentions}
\end{figure}

With respect to the number of mentions, Figure~\ref{fig:mentions} shows that viral tweets labelled as containing fake news appear to use mentions to other users less frequently than viral tweets not containing fake news. In other words, tweets containing fake news mostly contain 1 mention, whereas other tweets tend to have two). Such differences are statistically significant.

\begin{figure}[!ht]
 \centering
  \includegraphics[width=0.98\columnwidth, keepaspectratio]{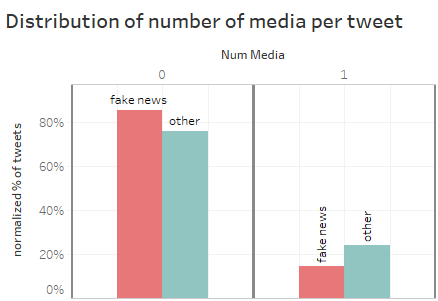}
  \caption{Number of media elements embedded within viral tweets (labelled as containing fake news vs. labelled as not containing them)}
  \label{fig:media}
\end{figure}

The analysis (Figure~\ref{fig:media}) of the  presence of media in the tweets in our dataset shows that tweets labelled as not containing fake news appear to present more media elements than those labelled as fake news. However, the difference is not statistically significant. 

\begin{figure}[!ht]
 \centering
  \includegraphics[width=0.98\columnwidth, keepaspectratio]{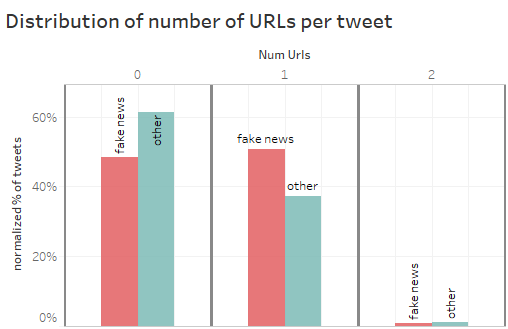}
  \caption{Number of URLs embedded within viral tweets (with fake news vs. without them). Differences are statistically significant with $\alpha = 0.05$}
  \label{fig:urls}
\end{figure}

On the other hand, Figure~\ref{fig:urls} shows that viral tweets containing fake news appear to include more URLs to other sites than viral tweets that do not contain fake news. In fact, the difference between the two distributions is statistically significant (assuming $\alpha = 0.05$).

\subsection*{Polarization}

Finally, manual inspection of the text field of those viral tweets labelled as containing fake news shows that 117 of such tweets expressed support for Donald Trump, while only 8 supported Hillary Clinton. The remaining tweets contained fake news related to other topics, not expressing support for any of the candidates.

\section*{Discussion}

As a summary, and constrained by our existing dataset, we made the following observations regarding differences between viral tweets labelled as containing fake news and viral tweets labelled as not containing them:
\begin{itemize}
\item Less than $0.1\%$ of the tweets went viral. Out of those, only $10\%$ were labelled as containing fake news. 
\item Tweets containing fake news that became viral during the day of the election were mostly created very shortly before that day or in the day. That contrasts with tweets not containing fake news (which were initially created much before election day).
\item Considering retweets, favourites and hashtags as proxies for exposures, we did not find any difference between viral tweets labelled as containing fake news and viral tweets labelled as not containing them. 
\item The characterization of accounts spreading fake news has shown that the proportion of unverified accounts that generates at least a tweet containing fake news is larger than that of accounts spreading tweets not labelled as fake news. 
\item Even if the accounts producing fake news are, on average, following the same number of other users than those producing tweet with no fake news in them, the distribution of followers are statistically different. 
\item There is no significant difference between the number of media elements in viral tweets labelled as containing fake news and viral tweets labelled as not containing them. 
\item Viral tweets labelled as containing fake news tend to have more URLs than viral tweets with labelled as not containing fake news. 
\item Regarding polarization, fake news were heavily supportive of the Trump campaign. 
\end{itemize}

{
\begin{table*}[htbp]
\begin{center}
\begin{tabular}{| >{}m{0.77\textwidth} | >{}m{0.19\textwidth} |} \hline
\multicolumn{2}{| c |} {Hypothesis}\\ \hline \hline
H1A: The average number of retweets of a viral tweet containing fake news is larger than that of viral tweets not containing them & NOT CONFIRMED \\ \hline
H1B: The average number of hashtags and user mentions in viral tweets with fake news is larger than that of viral tweets with no fake news in them. & NOT CONFIRMED \\ \hline
H2A: Viral tweets containing fake news have a larger number of URLs. & CONFIRMED \\ \hline
H2B: Creation date of an account generating tweets with fake news is more recent that those accounts tweeting non-fake news content. &  CONFIRMED \\ \hline
H2C: The rate of friends/followers of accounts tweeting fake news is larger than the rate of those creating tweets without them. & CONFIRMED \\ \hline
H3: Viral tweets containing fake news are slanted towards one candidate. & CONFIRMED\\ \hline
\end{tabular}
\caption{Summary of our conclusions, and tested hypothesis}
\label{table:hypothesis}
\end{center}
\end{table*}
}

These findings (related to our initial hypothesis in Table~\ref{table:hypothesis}) clearly suggest that there are specific pieces of meta-data about tweets that may allow the identification of fake news. One such parameter is the time of exposure. Viral tweets containing fake news are shorter-lived than those containing other type of content. This notion seems to resonate with our findings showing that a number of accounts spreading fake news have already been deleted or suspended by Twitter by the time of writing. If one considers that researchers using different data have found similar results~\cite{Allcot2017}, it appears that the lifetime of accounts, together with the age of the questioned viral content could be useful to identify fake news. In the light of this finding, accounts newly created should probably put under higher scrutiny than older ones. This in fact, would be a nice a-priori bias for a Bayesian classifier.

Accounts spreading fake news appear to have a larger proportion of friends/followers (i.e.\ they have, on average, the same number of friends but a smaller number of followers) than those spreading viral content only. Together with the fact that, on average, tweets containing fake news have more URLs than those spreading viral content, it is possible to hypothesize that, both, the ratio of friends/followers of the account producing a viral tweet and number of URLs contained in such a tweet could be useful to single-out fake news in Twitter. Not only that, but our finding related to the number of URLs is in line with intuitions behind the incentives to create fake news commonly found in the literature~\cite{Allcot2017} (in particular that of obtaining revenue through click-through advertising). 

Finally, it is interesting to notice that the content of viral fake news was highly polarized. This finding is also in line with those of Alcott et al.~\cite{Allcot2017}. This feature suggests that textual sentiment analysis of the content of tweets (as most researchers do), together with the above mentioned parameters from meta-data, may prove useful for identifying fake news.

\section*{Conclusions}
With the election of Donald Trump as President of the United States, the concept of \textit{fake news} has become a broadly-known phenomenon that is getting tremendous attention from governments and media companies. We have presented a preliminary study on the meta-data of a publicly available dataset of tweets that became viral during the day of the 2016 US presidential election. Our aim is to advance the understanding of which features might be characteristic of viral tweets containing fake news in comparison with viral tweets without fake news. 

We believe that the only way to automatically identify those deceitful tweets (i.e.\ containing fake news) is by actually understanding and modelling them. Only then, the automation of the processes of tagging and blocking these tweets can be successfully performed. In the same way that spam was fought, we anticipate fake news will suffer a similar evolution, with social platforms implementing tools to deal with them. With most works so far focusing on the actual content of the tweets, ours is a novel attempt from a different, but also complementary, angle. 

Within the used dataset, we found there are differences around exposure, characteristics of accounts spreading fake news and the tone of the content. Those findings suggest that it is indeed possible to model and automatically detect fake news. We plan to replicate and validate our experiments in an extended sample of tweets (until 4 months after the US election), and tests the predictive power of the features we found relevant within our sample.



\section*{Author Disclosure Statement}
No competing financial interest exist.

\bibliographystyle{bigdata} 
\bibliography{references}

\end{document}